\def\year{2022}\relax
\documentclass[letterpaper]{article} 
\usepackage{aaai22}  
\usepackage{times}  
\usepackage{helvet}  
\usepackage{courier}  
\usepackage[hyphens]{url}  
\usepackage{graphicx} 
\usepackage{natbib}  
\usepackage{caption} 
\DeclareCaptionStyle{ruled}{labelfont=normalfont,labelsep=colon,strut=off} 
\usepackage{algorithm}
\usepackage{algorithmic}

\usepackage{newfloat}
\usepackage{listings}
\floatstyle{ruled}
\newfloat{listing}{tb}{lst}{}
\floatname{listing}{Listing}
\usepackage{amsmath, amsthm, amsfonts} 
\usepackage{bm}
\usepackage{mathtools}
\usepackage{booktabs}
\usepackage{braket}
\usepackage{enumerate}
\usepackage{tikz} 
\usepackage{adjustbox}

\usepackage{ltablex}
\usepackage{adjustbox}
\usepackage{booktabs}
\usepackage[pass]{geometry}

\graphicspath{{images}}
\usetikzlibrary{arrows,shapes,positioning,shadows}

\DeclareMathOperator*{\argmax}{arg\,max}
\DeclareMathOperator*{\sign}{sign}

\DeclareMathOperator*{\sort}{sort}
\newcommand{\algnamn}{C2D}

\newtheorem{proposition}{Proposition}[section]
\newtheorem{lemma}{Lemma}[section]

\newtheorem*{repropobserve}{Proposition~\ref{thm:observeprop}}

\newtheorem*{repropconv}{Proposition~\ref{thm:convprop}}

\newtheorem*{relemma}{Lemma~\ref{thm:bellmanequi}}
\theoremstyle{definition}
\newtheorem{definition}{Definition}[section]

\title{Conjugated Discrete Distributions for Distributional Reinforcement Learning}
\author{Blind Review, Paper ID:8651}
\affiliations{Department of Mathematics, Linn\ae us University, Växjö, Sweden}
\author{
    Björn Lindenberg\thanks{
        \begin{tabular}{rl}
            Correspondence: &  bjorn.lindenberg@lnu.se \\
            Appendix: & https://github.com/bjliaa/c2d
        \end{tabular}}, Jonas Nordqvist, Karl-Olof Lindahl
}

\begin{document}

\maketitle

\begin{abstract}
  In this work we continue to build upon recent advances in reinforcement learning for finite Markov processes. A common approach among previous existing algorithms, both single-actor and distributed, is to either clip rewards or to apply a transformation method on Q-functions to handle a large variety of magnitudes in real discounted returns. We theoretically show that one of the most successful methods may not yield an optimal policy if we have a non-deterministic process. As a solution, we argue that distributional reinforcement learning lends itself to remedy this situation completely. By the introduction of a conjugated distributional operator we may handle a large class of transformations for real returns with guaranteed theoretical convergence. We propose an approximating single-actor algorithm based on this operator that trains agents directly on unaltered rewards using a proper distributional metric given by the Cramér distance. To evaluate its performance in a stochastic setting we train agents on a suite of 55 Atari 2600 games using sticky-actions and obtain state-of-the-art performance compared to other well-known algorithms in the Dopamine framework.
\end{abstract}

\section{Introduction}
In recent years, the resurgence of the distributional perspective in modern algorithms for reinforcement learning (RL) has represented a paradigm shift. Initially only used in risk-sensitive methods with parameterized continuous functions or in theoretical analysis, the approach of combining deep learning with distributional learning of return variables has been empirically proved to give superior performance. This is possibly due to a much richer set of predictions than value-based algorithms and different learning trajectories under gradient descent \cite{bellemare2017distributional, lyle2019comparative}. As such, \emph{distributional reinforcement learning} (DRL) is now the foundation of many state-of-the-art RL-algorithms such as the quantile functions of \cite{dabney2018implicit, dabney2018distributional, yang2019fully}. 
\subsubsection{Transformations}
Another prevalent method in modern algorithms is to shape or transform an underlying \emph{Markov decision process} (MDP) in order to improve performance. This is in itself an old idea in reinforcement learning. Transformations are particularly important for algorithms that are supposed to function over a wide variety of different environments with varying reward magnitudes, where adaptation to different settings is a common way to benchmark the robustness and ability of an algorithm. 

Ideally, one would want the fixed points of the Bellman operators in both the original MDP and the transformed MDP to induce the same optimal policy, \emph{i.e.}, to obtain \emph{optimal policy invariance}. Early attempts at shaping rewards directly where shown to require specific forms in terms of impractical potential functions if we were to keep the invariance \cite{rewardshaping}. A much more general approach is to consider non-linear equations that correspond to transformed Bellman operators. An investigation of such models and operator classes can be found in \cite{van2019general}. Regardless, many modern algorithms still employ reward shaping by clipping rewards for improved performance, even though this implies that the underlying MDP may never be solved in a theoretical sense. 

An attempt to learn on unaltered rewards was presented in \cite{observeandlook}, where an alternative Bellman operator $\mathcal{T}_h$ was introduced. The operator uses a function to generate updates with downscaled expectations. This allowed for deep Q-learning over environments with wildly different return magnitudes. At the time of writing, this method is used successfully by some of the most powerful known distributive algorithms \cite{r2d2, muzero, agent57}. However, the proof of invariance for $\mathcal{T}_h$ in \cite{observeandlook} required fully deterministic MDPs and it was left as an open question whether a similar property holds in a stochastic setting.

\subsubsection{A New Distributional Operator}
We show in our work that the previously mentioned invariance for $\mathcal{T}_h$ may not hold true if we are given a non-deterministic MDP. Specifically, we show that the fixed-point policy similarity may break for a class of operators which includes $\mathcal{T}_h$. We show that this problem is rectified in DRL by the introduction of a new generalized operator $T_\varphi$, the so-called \emph{conjugated distributional operator}. The operator, which transforms random outcomes by a chosen homeomorphism, can properly handle all its transformations while keeping the invariance property intact. 

\subsubsection{Our Algorithm}
To test the effects of training on unaltered rewards in DRL, we propose in the context of approximate DRL a single-actor algorithm \emph{conjugated discrete distributions } (\algnamn{}). The algorithm extends C51 in \cite{bellemare2017distributional} by using discrete distributions with freely moving supports and learns conjugate distributions by sampling $T_\varphi$. The extension still implies sampling done through distributional DQN, but instead of the fixed support and projection operator of C51, \algnamn{} uses parameterized embedded \emph{probability functions} and \emph{atom functions} to represent discrete measures. Moreover, since supports are no longer shared between measures, \algnamn{} replaces the cross-entropy loss of C51 by the squared \emph{Cramér distance}. This implies a DRL-algorithm that does minimization directly against a proper distributional metric. 

\subsubsection{Evaluation on Stochastic Games} The algorithm was evaluated on a suite of 55 Atari 2600 games where the use of sticky actions \cite{machado2018dopamine} induced non-deterministic MDPs. For an ``apples-to-apples" comparison with other algorithms we used the Dopamine framework protocol \cite{castro18dopamine}, where all involved agents were trained using a joint set of hyperparameters, including the sticky action probability. In the evaluation, \algnamn{} obtained state-of-the-art performance, comparable to Rainbow \cite{hessel2018rainbow}, with a significant performance improvement over C51. 

\subsubsection{Organization}
This paper is organized as follows. In Section~\ref{sec:setting} we state the foundations of DRL through the lens of measure theory, which is the established framework of \cite{rowland2018analysis}. In Section~\ref{sec:approach}, this allows for a much clearer picture of the connection between related work and our theoretical results in terms of the pushforward functions used to define our operator. We close Section~\ref{sec:approach} by discussing the three pillars of our C51 extension for approximate DRL. These are comprised of; \emph{distributional DQN} to sample operations, \emph{discrete measures} for distributional approximations and the \emph{Cramér distance} combined with a method for its computation. In subsequent sections we then explicitly present implementation details of \algnamn{} and present evaluation results on Atari with an ending discussion. A comprehensive view of graphs, data and parameters can be found in the appendix along with proofs.

\section{Setting}
\label{sec:setting}
Let $\mathcal{S}, \mathcal{A}$, and $\mathcal{R}$ be finite state, action and reward spaces. We consider an agent-environment interaction which produces random sequences 
\begin{equation*}
 S_0, A_0, R_1, S_1, A_1, R_1, \dots, S_t, A_t, R_{t+1}, S_{t+1}, A_{t+1}, \dots
\end{equation*}
according to the usual feedback loop of reinforcement learning. Our model for the interaction is a finite MDP $M = (\mathcal{S},\mathcal{A},\mathcal{R}, \rho)$, where $\rho$ is the transition kernel which maps $(s,a) \in \mathcal{S} \times \mathcal{A}$ to joint distributions $\rho(r, s' \mid s, a)$ over $\mathcal{R}\times \mathcal{S}$. In addition, the agent may sample its actions according to a stationary policy $\pi$ which maps states $s \in \mathcal{S}$ to distributions $\pi(s)$ over $\mathcal{A}$.

\subsection{Distributional Reinforcement Learning}
\label{sec:DRL}
Given a policy $\pi$ and a state-action pair $(s,a)$ at time $t$ we have a future discounted return in the form of the random variable
\begin{equation}
\label{eq:cumulative_reward}
    Z_{\pi}(s,a) = \left. \sum_{k = 0}^\infty \gamma^k R_{t + k + 1} \;\middle|\;S_t = s, A_t = a \right.,
\end{equation}  
where $\gamma \in (0,1)$ is a fixed discount factor.

Denote by~$\eta_{\pi}^{(s,a)} \in \mathcal{P}(\mathbb{R})$ the probability distribution of~$Z_{\pi}(s,a)$ and consider the \emph{collection} $\eta_{\pi}$ of all such probability distributions as the image of $(s,a) \mapsto \eta_{\pi}^{(s,a)}$. Then the expected return for any state-action pair can be found as the first moment
\begin{equation*}
    Q_{\pi}(s,a) \coloneqq \mathbb{E}\left[Z_{\pi}(s,a)\right] = \int_{\mathbb{R}} z \ d \eta_{\pi}^{(s,a)} (z),
\end{equation*}
\emph{i.e.}, the inherent expected value of being in some state $s$, taking action $a$ while following $\pi$.   

We note that any collection $\eta$ of probability distributions with bounded first moments defines a bounded Q-function, say $Q_{\eta}$ where $Q_{\eta}(s,a) \coloneqq \int_{\mathbb{R}} z \ d \eta^{(s,a)} (z)$. It follows that we can define a \emph{greedy} policy with respect to $\eta$ that selects any action $a^*$ in $\argmax_{a} Q_{\eta}(s,a)$.

Given a probability distribution $\mu \in \mathcal{P}(\mathbb{R})$ and a measurable function $f:\mathbb{R}\to \mathbb{R}$, the \emph{pushforward measure} $f\#\mu \in \mathcal{P}(\mathbb{R})$ is defined by 
$f \# \mu (A) \coloneqq \mu \left(f^{-1} (A) \right)$, for all Borel sets $A\subseteq \mathbb{R}.$ We may think of the pushforward measure $f\#\mu$ as obtained from $\mu$ by shifting the support of $\mu$ according to the map $f.$ When given a real reward $r$ and a probability measure $\mu$ of some random variable $Z$ we denote the pushforward measure arising from adding $r$ to the $\gamma$-discounted value of $Z$ by $f_{r,\gamma} \# \mu$, where $f_{r,\gamma}(z) \coloneqq r + \gamma z$. For brevity we denote any translation $r + z$ by $f_r$. 

The framework introduced in \cite{bellemare2017distributional, rowland2018analysis} defines the \emph{Bellman distributional optimality operator} $T^*$ on any $(s,a)$-indexed collection $\eta$ of probability measures by
\begin{equation}
\label{eq:optimal_operator}
 \left(T^*\eta\right)^{(s,a)} \coloneqq \int_{\mathcal{R}\times\mathcal{S}} f_{r,\gamma} \# \eta^{\left(s',a^*(s')\right)} \ d \rho(r,s' \mid s, a),
\end{equation}
where $a^*(s')$ is any action in $\argmax_{a'} Q_{\eta}(s',a')$. The mixture distribution of \eqref{eq:optimal_operator} can be seen as the "expected`` pushforward measure found while acting greedily with respect to $\eta$. It can be shown that $T^*$ in \eqref{eq:optimal_operator} is not a contraction map in any metric space of distributional collections \cite{bellemare2017distributional}. This is in contrast to the contraction map on bounded functions given by the \emph{Bellman optimality operator} 
\begin{equation}
    \label{eq:bellman_operator}
\left(\mathcal{T}^* Q\right)(s,a) \coloneqq \mathbb{E}_{\rho}\left[R + \gamma \max_{a'} Q(S',a') \;\middle|\; s , a\right].
\end{equation}
Regardless we have the following result.
\begin{lemma}
\label{thm:bellmanequi}
Let $M = (\mathcal{S}, \mathcal{A}, \mathcal{R}, \rho)$ be a finite MDP. If  $\eta$ is any collection with measures of bounded first moments, then the induced Q-function $Q_{T^* \eta}$ of $T^* \eta$ can be expressed as 
\begin{align*} 
Q_{T^* \eta}(s,a) &= \mathbb{E}_{\rho}\left[R + \gamma \max_{a'} Q_\eta(S',a') \;\middle|\; s, a\right] \\
&= \left(\mathcal{T}^* Q_\eta \right)(s,a)
\end{align*}
for all $(s,a) \in \mathcal{S} \times \mathcal{A}$.
\end{lemma}
\begin{proof}
In Appendix.
\end{proof}
Hence, the induced Q-function sequence $Q_k$ of $\eta_k \coloneqq \left(T^*\right)^k \eta_0$ satisfies $\forall \ (s,a)$,
\begin{equation}
\label{eq:momentbellman}
     Q_{k+1}(s,a) = \int_{\mathbb{R}} z \ d \eta_{k+1}^{(s,a)}(z) = \left(\mathcal{T}^* Q_k\right)(s,a).
\end{equation}
Indeed, given any starting collection $\eta_0$ with bounded moments we find that $Q_k$ converges to the \emph{optimal Q-function} $Q^*$ with respect to the uniform norm as $k \to \infty$, and generates an \emph{optimal policy} $\pi^*(s) \in \argmax_a Q^*(s,a)$ \cite{ bertsekas1996neuro, szepesvari2010algorithms, bellemare2017distributional, sutton2018reinforcement}.
\section{Our Approach and Related Work}
\label{sec:approach}
We begin this section by motivating and defining a generalized \emph{conjugated optimality operator} for DRL, which is key to our single-actor algorithm. We close the section by discussing our approach to approximate distributional reinforcement learning based on this operator.
\subsubsection{Value-based Transformation Methods}
\label{sec:conjop}
If we are training an agent which uses parameterized networks in an MDP that displays relatively high variance in both action-values and reward signals, then we may want to consider transformations of the involved quantities to improve stability. As an example, the original DQN-implementation in \cite{mnih2015human} clips rewards of varying orders into the interval $[-1,1]$. However, we note that this will encourage sequence-searching strategies instead of policies that can distinguish between actions that have relatively large differences in real returns. Thus, the clipping procedure may drastically change the underlying problem and its solution. An alternative approach presented in \cite{observeandlook} uses a procedure with unaltered rewards, which is well-suited for deterministic environments and is successfully used by several well-known algorithms \cite{r2d2, muzero, agent57}. The method scales with an invertible strictly increasing odd function,
\begin{equation}
\label{eq:observefunc}
    h(x) \coloneqq \sign(x) \left( \sqrt{1 + |x|} - 1 \right) + \epsilon x, \quad 0 < \epsilon \ll 1,
\end{equation}
which is (strictly) concave on $\mathbb{R}_+\coloneqq\set{x\in\mathbb{R} \mid x\geq 0}$. The operation $\mathcal{T}_h Q$ on Q-functions is for each $(s,a)$ then given by  
\begin{equation}
\label{eq:observe}
    \mathbb{E}_{\rho}\left[h\left(R + \gamma \max_{a'} (h^{-1} \circ Q)(S_{t+1},a') \right) \;\middle|\; s, a\right].
\end{equation}
However, finding a proper operator for transformations when working with Q-functions directly may not be straight forward. As correctly pointed out in \cite{observeandlook}, iteration by \eqref{eq:observe} may not yield optimal policy invariance in a stochastic setting. Explicitly we have the following result.
\begin{proposition}
\label{thm:observeprop}
Let $h$ be an invertible strictly increasing odd function which is strictly concave on $\mathbb{R}_+$ and define $\mathcal{T}_h$ by \eqref{eq:observe}. Then there exists a finite MDP where the fixed-point of $\mathcal{T}_h$ does not yield an optimal policy.
\end{proposition}
\begin{proof}
In Appendix.
\end{proof}
\subsubsection{Conjugated Optimality Operator}
The situation can be completely remedied in the setting of DRL by properly applying the conjugation in \eqref{eq:observe} on all outcomes, and by choosing actions that respect the transformation. More precisely, let $Z$, $W$ be random variables satisfying $W = h(Z)$. If we put $\mu_Z$ and $\mu_W$ as their respective distributions, then $\mu_Z = h^{-1}\#\mu_W$ and $\mu_W = h\#\mu_Z$. It follows that if $Z' \coloneqq r + \gamma Z$, then
\begin{equation*}
    W' \coloneqq h(Z') = h(r + \gamma Z) = h\left(r + \gamma h^{-1}(W) \right) 
\end{equation*}
obeys the law of $\left( h \circ f_{r, \gamma} \circ h^{-1}\right) \# \mu_W$. Moreover, for the expectation we obtain
\begin{equation*}
    \mathbb{E}\left[Z\right] = \int_\mathbb{R} z \ d \mu_Z = \int_{h(\mathbb{R})} h^{-1}(w) \ d \mu_W.
\end{equation*}
We are thus led to the following operator which works directly on collections of distributions for the transformed random variables.   
\begin{definition}
\label{def:operator}
Let $M = (\mathcal{S},\mathcal{A},\mathcal{R}, \rho)$ be a finite MDP, $J$ an open interval in $\mathbb{R}$ and $\varphi \colon \mathbb{R} \to J$ a homeomorphism. Furthermore, let $\xi \coloneqq \set{\xi^{\left(s, a\right)}}$ be a collection of probability measures over $J$, such that $\varphi^{-1}$ is integrable with respect to all measures in $\xi$. Put $g \coloneqq \varphi \circ f_{r, \gamma} \circ \varphi^{-1}$. Then the \emph{conjugated distributional optimality operator} $T_\varphi$ on collections $\xi$ is for~each~$(s,a)$ defined by
\begin{align*}
    \left(T_{\varphi} \xi\right)^{(s,a)}
    \coloneqq\int \displaylimits_{\mathcal{R} \times \mathcal{S}} g \# \xi^{\left(s', a^*\right)} \ d \rho(r, s' \mid s, a),
\end{align*}
where $a^*$ is chosen uniformly such that
\begin{equation*}
   \int_{J} \varphi^{-1}(w) \ d \xi^{(s',a')}(w) 
\end{equation*}
is maximized.
\end{definition}
Since $g = \varphi \circ f_{r, \gamma} \circ \varphi^{-1}$ will always be continuous, hence measurable, and since we focus on distributions where $\varphi^{-1}$ is integrable, $T_\varphi$ is well-defined. The fact that $T_\varphi$ now correctly mirrors iterations by $T^*$ in \eqref{eq:optimal_operator} is stated by the following result.
\begin{proposition}
\label{thm:convprop}
Let $\xi_0$ be an initial collection of measures on $J$ with supports contained in a closed bounded interval $I \subset J$. If we set
\begin{equation*}
   \xi_{k} \coloneqq T_{\varphi} \xi_{k-1} = T_\varphi^k \xi_0 
\end{equation*}
as the $k$th iteration of $\xi_0$ with respect to $T_\varphi$, then $Q_k$ defined by
\begin{equation*}
    Q_k(s,a) \coloneqq \int_{J} \varphi^{-1}(w) \ d \xi_k^{(s,a)}(w)
\end{equation*}
satisfies the Bellman iteration $Q_k = \mathcal{T}^* Q_{k-1}$.
\end{proposition}
\begin{proof}
In Appendix.
\end{proof}
\subsection{Approximate DRL}
\label{sec:approxtech}
In order to probe the possible benefits of learning in conjugate space by $T_\varphi$ in Definition~\ref{def:operator}, we present in this section the necessary approximation concepts needed for our single-actor algorithm. This includes the approximate operator method of C51 in \cite{bellemare2017distributional}, where we estimate any operation of $T_\varphi$ through a single observed transition. Moreover, compared to C51 we focus on representative measures taken from a larger class of discrete distributions where the aim is to train agents via a proper distributional metric.
\subsubsection{Distributional DQN}
\label{sec:distdqn}
The DQN-agent presented in \cite{mnih2015human} employs a deep Q-Network $Q(s,a;\bm{\theta})$ to approximate action values. During learning, an older periodically updated clone $Q(s,a;\bm{\theta}^-)$ is used as a stabilizing critic on current estimations. That is, given uniformly drawn transitions $(s,a,r,s')$ from a \emph{replay buffer} the learning algorithm of DQN approximates \eqref{eq:bellman_operator} by computing gradients on squared \emph{temporal difference errors} $\delta(s,a,r,s')^2$ where
\begin{equation*}
  \delta(s,a,r,s') \coloneqq r+ \gamma Q(s',a^*;\bm{\theta}^-) - Q(s,a;\bm{\theta}).
\end{equation*}
A similar approach for approximate DRL is through a parameterized distribution network $\xi(s,a;\bm{\theta})$. The network is coupled with an older copy $\xi(s,a;\bm{\theta}^-)$. Learning targets are taken as single sample approximations of $T_\varphi\xi(s,a;\bm{\theta}^-)$ and set to
\begin{equation}
\label{eq:target}
  \nu(r,s') \coloneqq \left(\varphi \circ f_{r,\gamma} \circ \varphi^{-1}\right)\# \eta(s',a^*; \bm{\theta}^-).
\end{equation}
Alternatively we can also choose the target $\Pi \nu(r,s')$ for some distributional projection $\Pi$. It follows that in order to extract gradients and push our current estimation towards the target we need a \emph{distributional temporal difference error}
\begin{equation}
\label{eq:distributional_temporal_difference}
  \Delta(s,a,r,s') \coloneqq  d \left(\nu(r,s'), \eta(s,a;\bm{\theta}) \right), 
\end{equation}
where $d$ is some divergence or distance on a suitable space of probability measures.

\subsubsection{Discrete Distributions}
In most real applications our distributional collections $\xi$ will need to be implemented by a parameterized function that outputs approximate distributions $\xi(s,a,\bm{\theta})$. The original approach of C51 approximated distributions with discrete measures on a fixed set of 51 atomic classes, \emph{i.e.}, \emph{categorical reinforcement learning}. A larger set of discrete measures can be considered if we use a fixed number of quantiles together with quantile regression, which is done in \cite{dabney2018distributional, dabney2018implicit} and with a fully parameterized quantile method in \cite{yang2019fully}. 

In our work we take a fully parameterized approach and focus on discrete measures of the form
\begin{equation*}
   \xi(s,a,\bm{\theta}) \coloneqq \sum_{i = 1}^N p_i \delta_{x_i}, 
\end{equation*}
where $\set{p_i}$ represents probabilities in the mixture distribution for the atomic classes $\set{x_i}$, and where $N$ is some predefined number of atoms to use in the approximation.

\subsubsection{Distributional Losses}
\begin{algorithm}[t]
\caption{Squared Cramér distance $\ell_2^2(\mu, \nu)$ for discrete distributions}
\label{alg:distance}
\begin{algorithmic}
\REQUIRE Distributions $\mu \coloneqq \sum_{i} p_i \delta_{x_i}$ and $\nu \coloneqq \sum_{j} q_j \delta_{y_j}$
\STATE \# Extended signed measure support
\STATE $(w_k) \coloneqq \sort(x_1, \dots, x_{n_\mu}, y_1, \dots, y_{n_\nu})$
\STATE $n \gets n_\mu + n_\nu$
\STATE $(\Delta w_k) \coloneqq (w_{k+1} - w_k)$, $\quad k = 1, 2, \dots, n - 1$
\FOR{$k=1$ \TO $n$}
    \STATE \# Signed measure mass
    \STATE $r_k \gets \left \{ \begin{array}{cc}
        p_i, & w_k = x_i \in (x_1, \dots, x_{n_\mu}),  \\
        -q_j, & w_k = y_j \in (y_1, \dots, y_{n_\nu}).
    \end{array} \right.$
    \STATE $P_k \gets \sum_{l=1}^k r_l$
\ENDFOR
\STATE $L \gets \sum_{k = 1}^{n - 1} P_k^2 \Delta w_k$
\ENSURE $L$
\end{algorithmic}
\end{algorithm}
The C51 algorithm employs parameterized distributions on a fixed support and implicitly use the \emph{Kullback-Liebler divergence} (KL) in \eqref{eq:distributional_temporal_difference} combined with a projection of $\nu(r, s')$ onto the support. However, it is well-known that KL is not a proper metric, requires a common support and has no concept of the underlying geometry. With varying supports it is then only natural to consider distances in \eqref{eq:distributional_temporal_difference} that measures similarity in outcomes instead of likelihoods, \emph{i.e.}, whose gradients induce probability mass transport while being sensitive to differences in outcomes between distributions. A common metric in this regard, used in both analysis and application, is the \emph{Wasserstein distance} which is connected to Kantorovich's formulation of optimal transport cost \cite{villani2008optimal, bellemare2017distributional}. The distance generates biased sample gradients and is hard to apply directly, but indirect applications together with quantile regression have been used to great effect in \cite{dabney2018distributional, dabney2018implicit, yang2019fully}. 

However, in this paper we will focus on another proper metric called the \emph{Cramér distance}, which in the univariate case squared is directly proportional to the \emph{statistical energy distance} \cite{rizzo2016energy}:
\begin{definition}
\label{def:cramer}
For probability measures $\mu, \nu$ in $\mathcal{P}(\mathbb{R})$ we define the \emph{Cramér distance} by
\begin{equation*}
    \ell_2(\mu, \nu) \coloneqq\left( \int_{\mathbb{R}} \left ( F_{\mu}(w) - F_{\nu}(w) \right )^2 \ d w \right)^{1/2},
\end{equation*}
where $F_{\mu}$, $F_{\nu}$ are the CDFs of each measure respectively.
\end{definition}
The Cramér distance has been successfully used in analysis of convergence properties of \eqref{eq:optimal_operator}, but also in modified form in real applications with linear approximations \cite{rowland2018analysis, bellemare2019distributional,lyle2019comparative}. As a proper metric it has a couple of attractive features that incorporates the underlying geometry: Like the Wasserstein distance it is \emph{translation invariant}, \emph{i.e.},  $\ell_2(f_r \# \mu, f_r \# \nu) = \ell_2(\mu, \nu)$, and \emph{scale sensitive} such that $       \ell_2\left((\gamma z)\# \mu, (\gamma z)\# \nu\right) = \sqrt{\gamma}\ell_2(\mu, \nu)$ for $\gamma > 0$. Additionally in the context of deep learning with parameterized networks, it has the desirable property of generating \emph{unbiased gradient estimators} when squared and used as a loss \cite{bellemare2017cramer}.

Compared to previous implementations of the Cramér distance for discrete distributions, the atomic classes we consider are no longer equally spaced and may vary freely via parameterization. Hence similar to quantile methods this will give us a larger set of representative measures, but where the added degrees of freedom will also demand a slightly more involved distance computation. The solution used in our work is summarized in Algorithm~\ref{alg:distance}, where we exploit the fact that $F_\mu - F_\nu$ represents the distribution function of the signed measure $\mu - \nu$.

\section{Our Learning Algorithm}
\label{sec:algoritm}
By combining the concepts in Section~\ref{sec:approach}, we propose in this section \emph{conjugated discrete distributions} (\algnamn{}) as an algorithm for approximate DRL using a chosen homeomorphism $\varphi$ for conjugate pushforwards. The algorithm uses parameterized networks $\xi(s,a;\bm{\theta})$ to represent discrete measures $(\mathbf{p}, \mathbf{x})\coloneqq\sum_{i=1}^N p_i \delta_{x_i}$ with $N$ number of atoms, and our chosen homeomorphism $\varphi$ dictates greedy actions according to $a^*$ in Definition~\ref{def:operator}. During training we maintain an older clone $\xi(s,a;\bm{\theta}^-)$ together with the target approximating technique of \eqref{eq:target} and we use Algorithm~\ref{alg:distance} to judge our current estimations. The learning algorithm is summarized in Algorithm~\ref{alg:learn}. 
\begin{algorithm}
\caption{Learning with \algnamn{}}
\label{alg:learn}
\begin{algorithmic}
\REQUIRE Homeomorphism $\varphi$, transition $(s,a,r,s')$
\STATE \# Current estimation
\STATE $\mu(\bm{\theta}) \coloneqq \xi(s,a;\bm{\theta})$
\STATE \# Chosen target
\STATE $\left(\mathbf{p}(a'), \mathbf{x}(a')\right) \coloneqq \xi(s',a';\bm{\theta}^-)$ for $a'$ in $\mathcal{A}$
\STATE $a^* \coloneqq \argmax_{a'} \sum_{i = 1}^N p_i(a') \varphi^{-1}\left(x_i(a')\right)$
\STATE $\left(\mathbf{p}, \mathbf{x}\right) \coloneqq \xi(s',a^*;\bm{\theta}^-)$
\STATE $\nu\coloneqq \left(\mathbf{p}, \varphi\left(r\mathbf{1} + \gamma \varphi^{-1}(\mathbf{x})\right)\right)$ (elementwise)
\STATE \# Squared Cramér distance 
\STATE $L(\bm{\theta}) \coloneqq \ell_2^2\left(\nu,\mu(\bm{\theta})\right)$
\ENSURE $\nabla_{\bm{\theta}} L\left(\bm{\theta}\right)$
\end{algorithmic}
\end{algorithm}

\subsection{Implementation Details}
\begin{figure}[t]
    \centering
    \includegraphics{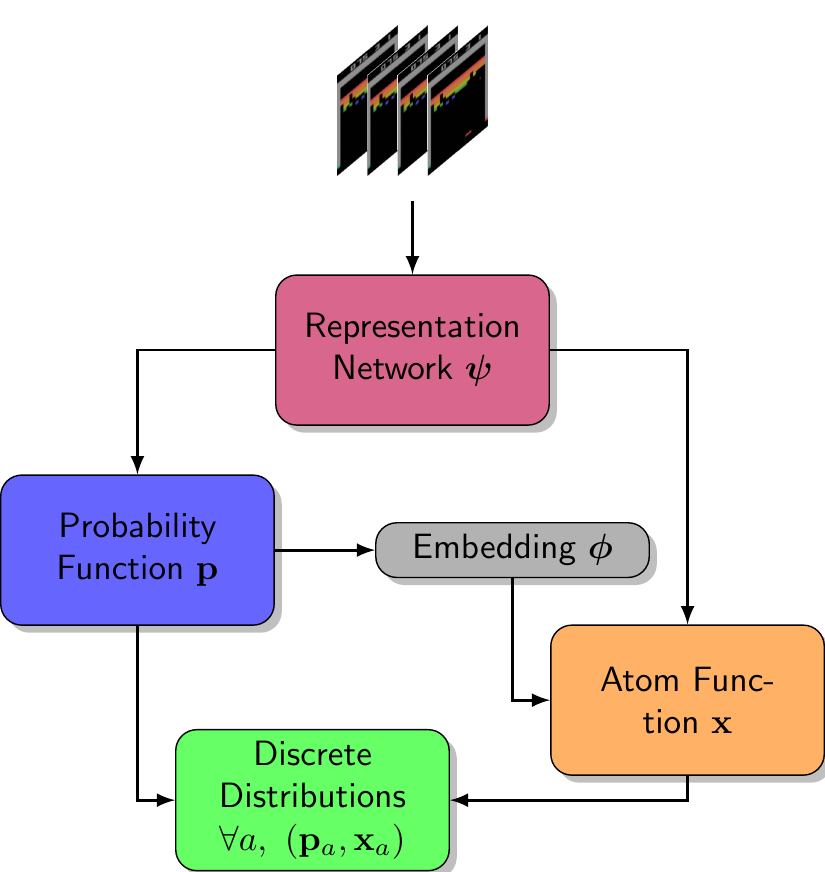}
    \caption{The functional model of \algnamn{} generates discrete distributions for all actions given an incoming state. A coupling effect between estimated probabilities $\mathbf{p}$ and atomic classes $\mathbf{x}$ is induced by a parameterized embedding $\phi$.}
    \label{fig:CADmodel}
\end{figure}
The functional model of \algnamn{}, shown in Figure~\ref{fig:CADmodel}, builds upon the representation network of DQN in \cite{mnih2015human}. An incoming state $s$ is encoded into a feature vector $\psi(s)$, which is then passed to a parameterized probability function $\mathbf{p}\left(\psi\right)$ and a parameterized atom function $\mathbf{x}\left(\phi, \psi \right)$ that jointly outputs discrete distributions for all actions. Similar to IQN and FQF in \cite{dabney2018implicit,yang2019fully} we employ an embedding $\phi$ of all probabilities to get a stabilizing coupling effect between probabilities and atomic classes. The embedding in \algnamn{} consists of an single dense layer with ReLU activation. The result vector $\phi(\mathbf{p})$ is concatenated with the feature vector $\psi(s)$ to yield an input for the atom function $\mathbf{x}\left(\phi, \psi \right)$. Moreover, since the supports only need to be ordered in distance computations we let the output of $\mathbf{x}$ vary unordered by the parameterization.

\subsubsection{Homeomorphism}
Our chosen homeomorphism $\varphi$ for the conjugated pushforward function in Algorithm~\ref{alg:learn} is a scaled variant of $h(x)$ in \eqref{eq:observefunc}. Namely, $\varphi(x) \coloneqq \beta h(x)$ where $\beta$ is close to but slightly less than 2. This is a less aggressive transformation than $h$ while keeping the invertible contraction property intact \cite{observeandlook}. Note that this implies $\varphi^{-1}(y) = h^{-1}(y/\beta)$ where $h^{-1}(x)$ is given by
\begin{equation*}
    \sign(x) \left( \left(\frac{\sqrt{1 + 4 \epsilon \left(|x| + 1 + \epsilon\right)}-1}{2\epsilon}\right)^2 - 1\right).
\end{equation*}
\subsubsection{Adaptive Scaling}
To make the model adaptable to varying orders of return magnitudes in different environments we use the activation
\begin{equation*}
    \alpha \tanh \left(x/c\right)
\end{equation*}
for all supports. Here $c$ is an implementation defined scaling hyperparameter for the output of the internal dense layer of $\mathbf{x}$, and $\alpha$ is a single trainable weight which is slowly tuned by gradients of the Cramér distance to properly accommodate for the scale of discounted returns in its present environment. 

\section{Evaluation on Stochastic Atari 2600 Games}
\label{s:results}
\begin{figure*}[t]
    \centering
    \includegraphics{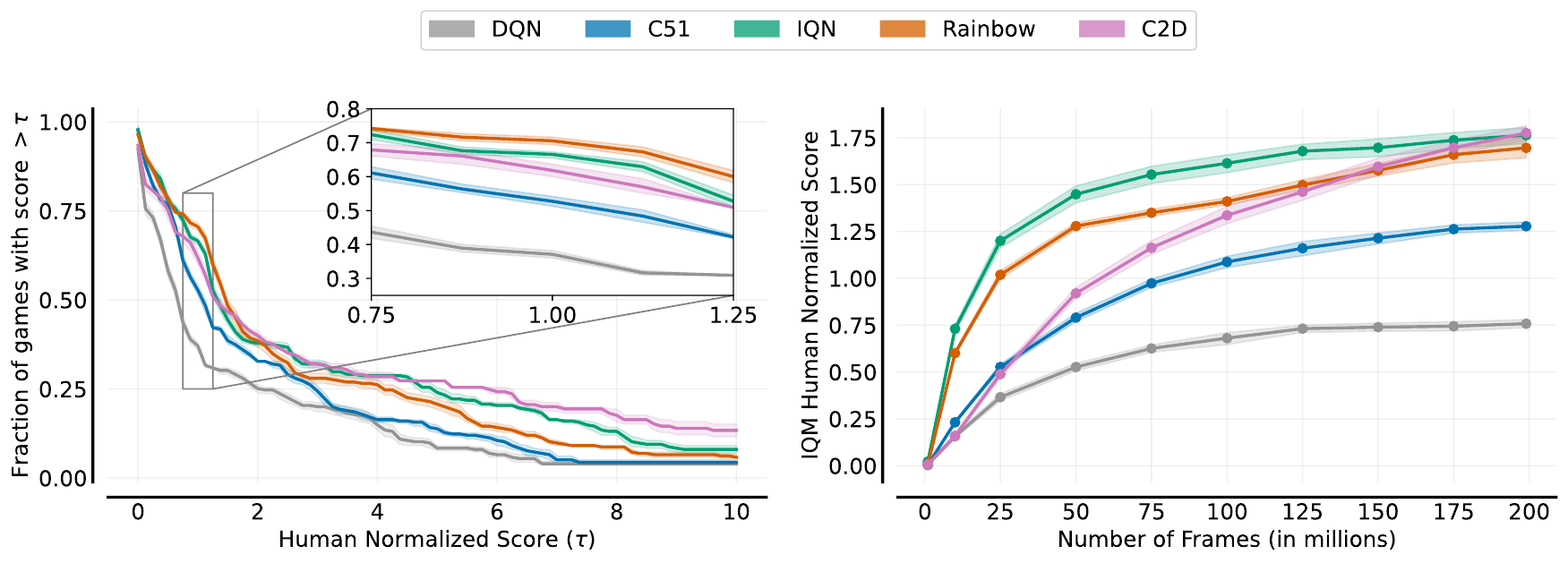}
    \caption{Accumulated statistics on 55 stochastic Atari games computed in accordance to the performance profiling methods given in \cite{agarwal2021deep}. The left shows the fraction of all games that after 200M frames achieve a score higher than threshold $\tau$. In particular, at $\tau = 1$ we see the fraction of games with human-like performance or better. The right shows training progression for the 25\% trimmed mean. Dopamine results are computed over 5 runs and \algnamn{} used 3 runs.}
    \label{fig:perfp}
\end{figure*}
In this section we gauge the performance of \algnamn{} by presenting our experiments on stochastic Atari 2600 games, where simulations are based upon the \emph{Arcade Learning Environment} framework \cite[ALE]{bellemare13arcade}. We instruct ALE to generate non-deterministic environments by using \emph{sticky actions} \cite{machado2018dopamine}. This implies that environments will have a non-zero probability of previous action repetition and this MDP altering effect is implemented internally in ALE.
\subsubsection{The Dopamine Protocol}
For an ``apples-to-apples" comparison we adopt the Dopamine protocol summarized in Table \ref{tab:hyppar} and evaluate training performance over a suite of 55 stochastic Atari 2600 games. The sticky action probability, in accordance to the protocol, is set to the default value of $0.25$ in ALE. As is common for single-actor algorithms evaluated on Atari, training procedures mostly follow the original implementation of DQN in \cite{mnih2015human}, which includes 1M-sized replay buffers, episode lengths capped at 108K frames (30 min) and a total training volume of 200M frames. There are however protocol specific settings such as the decay of $\varepsilon$-greedy action probabilities and the length of the random replay history. The protocol also dictates a fixed period for copying parameters to target networks. 
\begin{table}[h]
  \centering
  \begin{tabular}{lc}
    \toprule
    Parameter & Value \\
    \midrule
    Min. training $\varepsilon$ & 0.01 \\
    $\varepsilon$-decay schedule ($1.0 \to$ min. $\varepsilon$)  & 1M frames \\
    Min. history to start learning  & 80k frames \\
    Target network update frequency & 32k frames \\
    Sticky actions & 0.25 \\
    \bottomrule
  \end{tabular}
  \caption{Hyperparameters common to all implementations of the Dopamine protocol, where $\varepsilon$ dictates probabilities for explorative $\varepsilon$-greedy actions.}
  \label{tab:hyppar}
\end{table}
\subsubsection{\algnamn{} Specific Settings}
When applicable we set the hyperparameters of \algnamn{} as close as possible to other comparable algorithms. This includes setting $N = 32$ for the number of atomic classes used by the distributional approximations, which can be compared to the size of the quantile fraction set in IQN. It also includes using \emph{ADAM} as the network optimizer with learning rate $0.5\cdot10^{-4}$ and epsilon value $3.125\cdot10^{-4}$ \cite{adam}. 

Moreover, the padding in the three convolutional layers of the representation network $\psi$ follows that of DQN, which implies substantially less network weights than similar implementations found in Dopamine. However, our function implementation of $\psi$ will still differ from DQN by the incorporation of batch normalization layers between convolutions \cite{batchnorm}. We found that this makes specialized weight initialization techniques for the deep network to become largely redundant. 
 
We performed a preliminary search for \algnamn{}-specific parameters by measuring early training progression over six different environments with varying reward magnitudes. This resulted in using $\beta = 1.99$ for our transformation homeomorphism $\varphi(x) = \beta h(x)$. It also resulted in using $\alpha = 50$ and $c = 5$ for the support activation $\alpha \tanh \left(x/c\right)$ in each game. This implies that our \emph{initial distributions} will have a maximum support interval $(-50,50)$ to represent transformed outcomes in discounted returns. However since $\alpha$ is a trainable parameter, this maximum interval will either shrink or grow depending on both the environment and agent performance.
\begin{figure*}[t]
    \centering
    \includegraphics{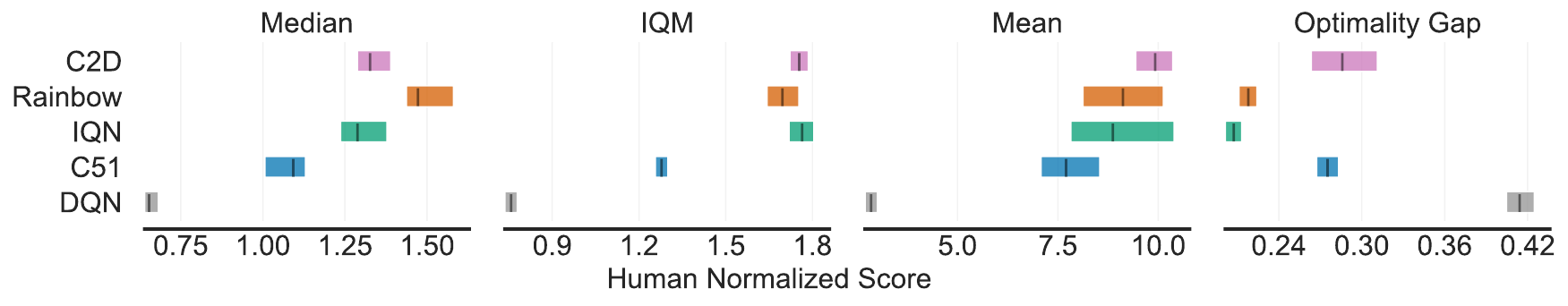}
    \caption{Aggregate metrics on Atari-200M over 55 games. The metrics are computed in accordance to the performance profiling methods given in \cite{agarwal2021deep}. Dopamine results are computed over 5 runs and \algnamn{} used 3 runs.}
    \label{fig:mean_med}
\end{figure*}
\subsubsection{Scores and Baselines}
Under a similar set of hyperparameters, we compare and evaluate training performance in terms of human-normalized scores \cite{mnih2015human}. With three runs for each game, comparisons are done against other sticky-action algorithms found in the Dopamine framework \cite{castro18dopamine}. The discrete measures of C51 \cite{bellemare2017distributional} is of interest since we can get a direct comparison of fixed versus varying support. In addition, we may observe the effect of using a proper distributional metric such as the Cramér distance as a loss versus the derived cross-entropy of KL. For comparisons with other strong distributional algorithms, we include IQN \cite{dabney2018implicit}, which is based on quantile regression, and Rainbow \cite{hessel2018rainbow}. Rainbow, with its myriad of techniques, is built upon C51 and is at the time of writing state-of-art within the Dopamine framework. It is of particular interest to us since performing close to Rainbow suggests that further improvements and superior performance can easily be obtained by adding any of the additional flavors found in Rainbow's non-distributional methods. Finally, as is the norm in most Atari 2600 evaluations, we also include DQN as a baseline.
\subsection{Results}
\begin{figure}[t]
    \centering
    \includegraphics{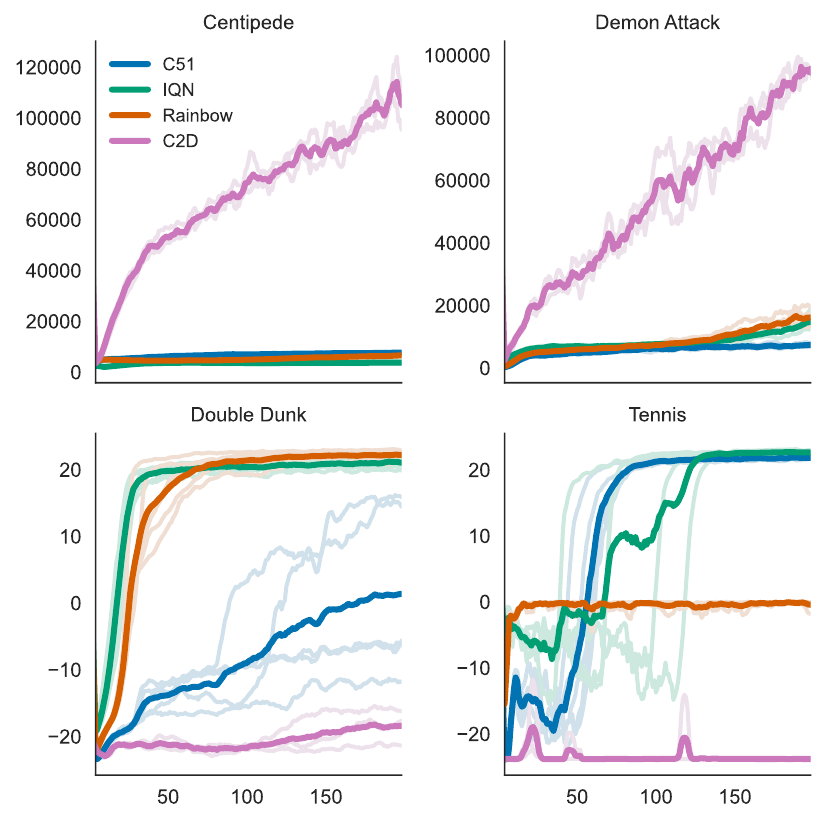}
    \caption{Moving averages (5M frames) of mean raw scores in four environments. With its stated settings, \algnamn{} excels in games with frequent rewards of relatively high variance and magnitude. However, \algnamn{} is often slow to learn or sensitive to exploration in environments with sparse rewards of low magnitude.}
    \label{fig:strongweak}
\end{figure}

In order to have a rigorous evaluation methodology, we analyze results by using the profiling tools presented in \cite{agarwal2021deep}. In Figure~\ref{fig:perfp} we see the fraction of all games that achieve a human-normalized score higher than threshold $\tau$. We also see the 25\% trimmed mean~(IQM) training progression over all frames. In Figure~\ref{fig:mean_med} we show 200M aggregate metrics measured in human-normalized scores over all 55 Atari games. These include the optimality gap, which measures the complement $1-x$ of the mean $x$ given that all scores are truncated by the human score. Thus, an optimality gap of 0 would indicate at least human-like performance over all games. To further showcase possible strengths and weaknesses, we present in Figure~\ref{fig:strongweak} four examples of mean training progressions, where \algnamn{} with its stated settings displayed significantly different trajectories compared to the other algorithms. More training graphs, support evolution and cross sections in raw scores is provided in the appendix. 

The IQM training progression in Figure~\ref{fig:perfp} indicates that \algnamn{} may achieve better long-term performance in this metric than the other baselines. This is also indicated in the overall mean at 200M in Figure~\ref{fig:mean_med}, which is heavily weighed by scores in environments where RL-algorithms to a substantial degree outperform humans. We can see this reflected in Figure~\ref{fig:perfp} where \algnamn{} maintains super-human performance in a significant portion of games ($\tau > 5$). However we note that the cross-section $\tau = 1$ indicates that \algnamn{} performs worse than IQN and Rainbow when it comes to number of games with human-like performance. In particular, the optimality gap and Figure~\ref{fig:strongweak} suggests that the algorithm may in the mean perform worse than C51 in games where it is weak. Moreover, the algorithm has a relatively slow initial progression, requiring more than 150M frames before it has comparable performance to IQN and Rainbow in the trimmed mean metric. In the median we find that \algnamn{} is on par with IQN.
\section{Discussion}
We have shown that previous value-based methods for learning with a transformed Q-function operator may not hold in a stochastic setting. As a theoretical solution in DRL, we have introduced and shown that the generalized operator $T_\varphi$ of Definition~\ref{def:operator} properly reflects learning in a space of distributions that is derived from a chosen transformation homeomorphism. We have also proposed an approximate DRL algorithm based on $T_\varphi$ in \algnamn{}, which has been evaluated on a suite of stochastic Atari games. The learning algorithm produced state-of-the-art performance with respect to the strongest single-actor algorithms in the Dopamine framework.  Specifically given the available seeds, we found that \algnamn{} outperformed both IQN and Rainbow in the averaged mean over all environments. In the median and IQM we found that \algnamn{} has IQN-comparable performance. We could also infer from the optimality gap that \algnamn{} is generally weaker than C51 in its low-performing environments.

Since $T_\varphi$ induces guaranteed theoretical Q-function convergence we argue that the operator serves as a sound basis for approximate DRL involving unaltered rewards. Given a stochastic setting, it seems hard to define a practical and fully transformation correct operator for value-based learning, \emph{i.e.}, a Q-function operator which handles transformations of the underlying random variables and where we have optimal policy invariance. For a deeper look into this problem, see \cite{van2019general}. However, the ease at which we may handle transformations through DRL is further indicative of the strength of the distributional formalism versus its classical counterpart.

A possible explanation for the success of \algnamn{} in the mean is that environments that heavily skew this statistic have frequent non-zero rewards with high variance in discounted returns. That is, the required conjugate distributional statistics for agent networks are then frequently updated by the Cramér distance, with large disparities between action-values in real discounted outcomes. In comparison and directly related to the IQM statistic, \algnamn{} often has significantly worse learning progression than IQN and sometimes subpar performance in environments with sparse rewards and low variance in returns, with a tendency to get stuck in non-optimal regions due to the inherit poor exploration of $\varepsilon$-greedy actions. Slowly updating statistics by moving probabilities and supports via the Cramér distance seems to be a harder optimization problem, sensitive to exploration. Although similar to the reasoning made in \cite{dabney2018implicit}, we argue that a rainbowesque version of \algnamn{} should easily obtain superior performance by adding from Rainbow's non-distributional arsenal, which contains techniques such as multi-step sampling, noisy networks and double distributional DQN. 

Interesting future directions for exploring the full scope of approximate DRL derived from $T_\varphi$ would be through the explorative and distributive methods of algorithms such as R2D2, MuZero or Agent57 \cite{r2d2, muzero, agent57}. This could include: A significant increase in training frames to observe long-term convergence behavior combined with an ablation study of the effects of proper transformations. Better representation functions through recurrent or residual networks. Better operational approximations through planning, multi-step sampling or full episode sampling. Better exploration methods for statistics gathering in order to avoid premature convergence on non-optimal greedy regions in difficult environments.

\section*{Acknowledgments}
We would like to thank the anonymous referees for their excellent feedback and questions. We would also like thank Morgan Ericsson, Department of Computer Science and Media
Technology, Linn\ae us University, for his technical assistance with the LNU-DISA High Performance Computing Platform.


\bibliography{ref}
\clearpage
\newgeometry{textwidth=6in}
\onecolumn
\appendix
\section{Mathematical Details and Proofs}
In this section we provide the proofs of Proposition~\ref{thm:observeprop} and Proposition~\ref{thm:convprop}. The propositions are restated before each proof for convenience. 
\begin{repropobserve}
Let $h$ be an invertible strictly increasing odd function which is strictly concave on $\mathbb{R}_+$ and define $\mathcal{T}_h$ by \eqref{eq:observe}. Then there exists a finite MDP where the fixed-point of $\mathcal{T}_h$ does not yield an optimal policy.
\end{repropobserve}
\begin{proof}
We recall from \eqref{eq:observe} that the operator $\mathcal{T}_h$ is given by
\begin{equation}
\label{eq:reobserve}
    (\mathcal{T}_h Q) (s,a) \coloneqq \mathbb{E}_{\rho}\left[h\left(R_{t+1} + \gamma \max_{a'} (h^{-1} \circ Q)(S_{t+1},a') \right) \;\middle|\; S_t = s, A_t = a\right].
\end{equation}
Hence an induced policy selects actions by $\argmax_a h^{-1}\left(Q(s,a)\right) = \argmax_a Q(s,a)$. We will now show that the optimal policy invariance property of $\mathcal{T}_h$ might break whenever the environment gives us a choice between a fully deterministic path and a path where the return variable is of maximum possible variance. 

Let $M = (\mathcal{S}, \mathcal{A}, \mathcal{R}, \rho)$ be a finite MDP with a non-terminal state $s$ and suppose that $\mathcal{A}$ consists of two actions $a, b$. Suppose further that $\mathcal{R}$ and $\rho$ are defined as follows: If we choose $a$ in $s$, then $\rho$ may send us to two different terminal states with equal probability, where we observe rewards $0$ and $R > 0$ respectively. On the other hand if we select $b$, then $\rho$ will send us deterministically to a single terminal state where we observe a reward $r$, which is defined by
\begin{equation}
\label{eq:simple}
 h^{-1}\left(\frac{h(R)}{2} \right) < r < \frac{R}{2}.
\end{equation}
To motivate the existence of such an $r$, note that $h^{-1}$ is strictly convex on $\mathbb{R}_+$ with $h^{-1}(0) = 0$, since, by the condition of the proposition, $h$ is strictly concave on $\mathbb{R}_+$ with $h(0) = 0$. Thus from Jensen's inequality we have
\begin{equation*}
    h^{-1}\left(\frac{h(R)}{2} \right) = h^{-1}\left(\frac{h(0)}{2} + \frac{h(R)}{2} \right) < \frac{h^{-1}\left(h(0)\right)}{2} +  \frac{h^{-1}\left(h(R)\right)}{2} = R/2.
\end{equation*}
It is clear that an optimal policy chooses the non-deterministic path of $a$ since the expected value is
\begin{equation*}
    0/2 + R/2 = R/2 > r. 
\end{equation*}
However, in the transformed approach of \eqref{eq:reobserve} we find the expectation of each action in the values
\begin{equation*}
    h(R)/2 + h(0)/2 = h(R)/2
\end{equation*}
for $a$ and $h(r)$ for $b$. Thus, given the way that $r$ was defined in \eqref{eq:simple} we have
\begin{equation*}
    h^{-1}(h(r)) = r > h^{-1}\left(h(R)/2 \right).
\end{equation*}
It follows that the fixed point inducing policy of \eqref{eq:reobserve}, found by one iteration, suggests that we choose non-optimal $b \neq a$, which completes the proof of the proposition.
\end{proof}
The following lemma is used in the proof of Proposition~\ref{thm:convprop} which is stated and proved directly after the lemma. 
\begin{relemma}
Let $M = (\mathcal{S}, \mathcal{A}, \mathcal{R}, \rho)$ be a finite MDP. If  $\eta$ is any collection with measures of bounded first moments, then the induced Q-function $Q_{T^* \eta}$ of $T^* \eta$ can be expressed as 
\begin{equation*} 
Q_{T^* \eta}(s,a) = \mathbb{E}_{\rho}\left[R + \gamma \max_{a'} Q_\eta(S',a') \;\middle|\; s, a\right] = \left(\mathcal{T}^* Q_\eta \right)(s,a)
\end{equation*}
for all $(s,a) \in \mathcal{S} \times \mathcal{A}$.
\end{relemma}

\begin{proof}
The result follows from a simple reformulation by the finiteness of the MDP, linearity of integration over mixture distributions and a change of variables
\begin{align*}
Q_{T^* \eta}(s,a) &= \int_{\mathbb{R}} z \ d \left(T^* \eta\right)^{(s,a)}
      =\int_{\mathbb{R}} z \ d \left(  \int_{\mathcal{R}\times\mathcal{S}} f_{r,\gamma} \# \eta^{\left(s',a^*\right)} \ d \rho(r,s' \mid s, a) \right) (z) \nonumber\\
    &= \int_{\mathcal{R} \times \mathcal{S}} \left( \int_{\mathbb{R}} z \ d \left( f_{r,\gamma} \# \eta^{\left(s',a^*\right)}  \right) (z) \right) \ d \rho(r,s' \mid s, a) \nonumber \\
    &= \int_{\mathcal{R} \times \mathcal{S}} \left( \int_{\mathbb{R}} \left(r + \gamma z \right)\ d \eta^{\left(s',a^*\right)} (z) \right) \ d \rho(r,s' \mid s, a) \nonumber\\   
    &=\mathbb{E}_{\rho}\left[R + \gamma \max_{a'} Q_\eta(S',a') \;\middle|\; s, a\right] = \left(\mathcal{T}^* Q_\eta \right)(s,a).  
\end{align*}
This completes the proof of the lemma.
\end{proof}

\begin{repropconv}
Let $\xi_0$ be an initial collection of measures on $J$ with supports contained in a closed bounded interval $I \subset J$. If we set
\begin{equation*}
   \xi_{k} \coloneqq T_{\varphi} \xi_{k-1} = T_\varphi^k \xi_0 
\end{equation*}
as the $k$th iteration of $\xi_0$ with respect to $T_\varphi$, then $Q_k$ defined by
\begin{equation*}
    Q_k(s,a) \coloneqq \int_{J} \varphi^{-1}(w) \ d \xi_k^{(s,a)}(w)
\end{equation*}
satisfies the Bellman iteration $Q_k = \mathcal{T}^* Q_{k-1}$.
\end{repropconv}

\begin{proof}
We recall from Definition~\ref{def:operator} that 
\begin{equation*}
  \left(T_{\varphi} \xi\right)^{(s,a)}  \coloneqq \int_{\mathcal{R} \times \mathcal{S}} \left(\varphi \circ f_{r, \gamma} \circ \varphi^{-1} \right) \# \xi^{\left(s', a^*\right)} \ d \rho(r, s' \mid s, a),
\end{equation*}
where $a^* = a^*(s')$ is chosen according to
\begin{equation*}
    a^*(s') \in \set{\argmax_{a'} \int_{J} \varphi^{-1}(w) \ d \xi^{(s',a')}(w)}.
\end{equation*}
Since $\varphi \circ f_{r, \gamma} \circ \varphi^{-1}$ is continuous, hence measurable, and $\varphi^{-1}$ by definition is integrable with respect to all involved measures, $T_\varphi$ is well-defined. Moreover, given any collection $\zeta$ over $J$ or $\mathbb{R}$ and any measurable function $f \colon J \to \mathbb{R}$ or $f \colon \mathbb{R} \to J$, we define $f \# \zeta \coloneqq \set{f\# \zeta^{(s,a)}}$ as a joint push-forward. 

Put $\eta_k \coloneqq \varphi^{-1} \# \xi_k$ and note that $\xi_k = \varphi \# \eta_k$ for all $k$. This implies $    \eta_k = \varphi^{-1}\#T_\varphi \left(\varphi \# \eta_{k-1}\right)$. Since we are dealing with finite MDPs, any integral over $\mathcal{R}\times\mathcal{S}$ with respect to $\rho$ given $(s,a)$ can be represented as a finite sum. So from the linearity of push-forwards and finite sums we have
\begin{align*}
 &\eta_k^{(s,a)} = \varphi^{-1}\# \int_{\mathcal{R}\times\mathcal{S}} \left(\varphi \circ f_{r,\gamma} \circ \varphi^{-1}\right) \# \left(\varphi \# \eta_{k-1} \right)^{(s', a^*)} \ d \rho(r, s' \mid s,a) \\
 &= \left(\varphi^{-1} \circ \varphi \right)\# \int_{\mathcal{R}\times\mathcal{S}} f_{r,\gamma} \# \eta_{k-1}^{(s', a^*)} \ d \rho(r, s' \mid s,a) = \int_{\mathcal{R}\times\mathcal{S}} f_{r,\gamma} \# \eta_{k-1}^{(s', a^*)} \ d \rho(r, s' \mid s,a).
\end{align*}
Moreover, by a change of variables and the assumption that $\varphi^{-1}$ is integrable, we find
\begin{align*}
    &\int_{J} \varphi^{-1}(w) \ d \xi_k^{(s,a)}(w) =     \int_{J} \varphi^{-1}(w) \ d \left(\varphi \# \eta_k^{(s,a)}\right)(w)
    =\int_{\mathbb{R}} z \ d \eta_k^{(s,a)}(z).
\end{align*} 
Hence $\eta_k$ corresponds precisely to iterations of the DRL optimality operator in \eqref{eq:optimal_operator}, \emph{i.e.}, $\eta_k = T^* \eta_{k-1}$ with an initial collection $\eta_0 = \varphi^{-1} \# \xi_0$. In particular, the induced Q-function sequence $Q_k$ of $\xi_k$ equals
\begin{equation*}
    Q_k(s,a) := \int_{J} \varphi^{-1}(w) \ d \xi_k^{(s,a)}(w) = \int_{\mathbb{R}} z \ d \eta_k^{(s,a)}(z) = Q_{\eta_k}(s,a).
\end{equation*}
Thus by Lemma~\ref{thm:bellmanequi}, we find
\begin{align*}
    Q_k(s,a) = Q_{\eta_{k}}(s,a) = Q_{T^* \eta_{k-1}}(s,a) = \left(\mathcal{T}^* Q_{\eta_{k-1}}\right)(s,a) = \left(\mathcal{T}^* Q_{k-1}\right)(s,a),
\end{align*}
which implies that they are iterates of the Bellman operator in \eqref{eq:bellman_operator}. It is now well-known from classical theory that since our MDP is finite with discount $\gamma < 1$ and $\xi_0$ induces a bounded Q-function $Q_0$, the subsequent iterates $Q_k$ are bounded and will converge uniformly to the optimal value function $Q^*$ as $k \to \infty$ \cite{szepesvari2010algorithms}.
\end{proof}
\section{Atari MDPs, Architecture and Hyperparameters}
In this section we present implementation details of our \algnamn{}-Atari experiments. This includes the computational details for our networks and values for used hyperparameters.
The Atari implementation of \algnamn{} used the ALE C++ library for simulations by encapsulating ALE in an environment class, which also handled storage of observed transitions in a circular replay buffer. Network computations and training was done in Python with TensorFlow 2.X by using a thin wrapper for the data exchange with C++.
\subsection{Atari MDPs}
Our Atari 2600 MDPs were induced by the settings in Table~\ref{tab:ale}. Following DQN we repeat each action 4 times in ALE and represent an observation by max-pooling ALE screens \#3 and \#4 in the generated sequence of 4 grayscaled screens. This is done in order to remove flicker due to partial screen updates. Observation frames, rescaled to $84 \times 84$ pixels, are then stacked 4 times and rolled by each step taken by an agent to form states of tensor dimensions $(4,84,84)$. Thus, states now represent short temporal views of game dynamics. To induce non-determinism we use sticky actions, which is handled internally in ALE. In addition, every episode is terminated after roughly 30min (108k frames), which is the default termination time for single-actor algorithms.
\begin{table}[h]
  \centering
  \begin{tabular}{ll}
    \toprule
    \textbf{Parameter} & \textbf{Value} \\
    \midrule 
    ALE version & 6.2 \\
ALE color spectrum & Grayscaled \\
ALE frame dimensions & $84 \times 84$ \\
Max episode length & 27k steps (108k frames) \\
Action repetition & 4 \\
State observation stacking & 4 \\
Terminal on life loss & True \\
Sticky actions & 0.25 \\
Discount $\gamma$ & 0.99 \\
\bottomrule
  \end{tabular}
  \caption{Atari specific settings.   \label{tab:ale}}
\end{table}
\subsection{Architecture}
The overall architecture of Figure~\ref{fig:CADmodel} for the Atari implementation follows that of DQN. States are represented by a sequence of 4 max-pooled observations and given to an encoding function $\psi$ consisting of three convolutional layers, interleaved by batch normalization + ReLU, and a ReLU-activated dense layer which computes a 512-feature sized vector $\psi(s)$. The encoded state $\psi(s)$ is then passed to a probability network $\mathbf{p}$ that computes $|\mathcal{A} \times N|$ probabilities $\mathbf{p}\left(\psi(s)\right)$ through a dense layer with softmax activation. The probabilities are also passed to a dense embedding layer $\phi$ which computes a 512-sized vector $\mathbf{e}\coloneqq \phi\left(\mathbf{p}\left(\psi(s)\right)\right)$, again with ReLU-activation. The embedding $\mathbf{e}$ and the feature vector $\psi(s)$ are concatenated to form an 1024-input vector, which is fed to an atom network $\mathbf{x}$. The resulting atoms $\mathbf{z}(\mathbf{e}, \psi(s))$ are computed by a dense layer of $|\mathcal{A} \times N|$ units with activation $\alpha \tanh(x/c)$. The combined output of the network is $(\mathbf{p}, \mathbf{x})$ which represents our discrete distributions, one for each available action at the current state $s$.
\subsection{\algnamn{} Settings}
Finally, Table~\ref{tab:sett} lists all other settings and hyperparameters used by \algnamn{} in our experiments. 
\begin{tabularx}{\linewidth}{@{}ll@{}}
\toprule
\textbf{Parameter} & \textbf{Value} \\
\midrule
\endfirsthead
\toprule
\textbf{Parameter} & \textbf{Value} \\
\midrule
\endhead
\midrule
\multicolumn{2}{r}{\footnotesize( To be continued)}
\endfoot
\bottomrule
\\\caption{\algnamn{} settings for our Atari experiments.\label{tab:sett}}
\endlastfoot
    TensorFlow version & 2.5 \\
    Number of atoms $N$ (IQN) & 32 \\
    Optimizer (IQN) & ADAM \\
    Learning rate (IQN) & $0.5\cdot10^{-4}$ \\
    ADAM epsilon (IQN) & $3.125\cdot10^{-4}$ \\
    ADAM global clip norm & 10.0 \\
    Training volume (DQN) & 50M steps (200M frames) \\
    Replay buffer (DQN) & 1M transitions $(s,a,r,s')$\\
    Random history (Dopamine) & 20k steps (80k frames) \\
    Initial training $\varepsilon$ (Dopamine) & 1.0 \\
    Minimum training $\varepsilon$ (Dopamine) & 0.01 \\
    $\varepsilon$-decay schedule ($1.0 \to$ min. $\varepsilon$) (Dopamine) & 250k steps (1M frames) \\
    Target network update frequency (Dopamine) & 8k steps (32k frames) \\
    Training frequency (DQN) & Every 4th step \\
    Batch size (Dopamine) & 32 buffered transitions $(s,a,r,s')$, uniformly sampled. \\
    Loss & Cramér distance $\int (F_\mu - F_\nu)^2 \ d w$ \\
    $h(x)$ & $\sign(x) \left( \left(\sqrt{1 + |x|} - 1 \right) + \epsilon x\right) $, $\epsilon = 0.001$ \\
    $h^{-1}(x)$ &  $\sign(x) \left( \left(\frac{\sqrt{1 + 4 \epsilon \left(|x| + 1 + \epsilon\right)}-1}{2\epsilon}\right)^2 - 1\right)$, $\epsilon = 0.001$\\
    Transformation scaling $\beta$ & 1.99 \\
    Homeomorphism $\varphi(x)$ & $\beta h(x)$ \\
    Inverse $\varphi^{-1}(x)$ & $h^{-1}(x/\beta)$ \\
    Support scale initialization & $ \alpha = 50.0$ (trainable) \\
    Internal output scaling $c$ & 5.0 \\
    Atom activation function & $\alpha \tanh{x/c}$ \\
\end{tabularx}
\newgeometry{onecolumn, total={8.5in,11in},top=0.75in, left=0.75in, right=0.75in}
\section{Atari Mean Scores for \algnamn{} (Sticky Action)}
\vspace{1em}
\begin{table}[ht]
    \vskip -1em
    \centering
    \small 
    \begin{tabular}{lcccc}
    \toprule
    Game & 10M & 50M & 100M & 200M\\
    \midrule
alien&613.0 (12.0)&1497.1 (214.7)&2826.9 (253.2)&4111.3 (340.8)\\
amidar&94.9 (1.7)&453.2 (25.3)&664.9 (77.1)&816.4 (68.3)\\
assault&2170.6 (297.4)&3162.6 (205.0)&3819.0 (541.3)&4997.0 (508.2)\\
asterix&2075.1 (191.4)&9818.6 (533.3)&19882.2 (3322.0)&62677.6 (8152.6)\\
asteroids&762.6 (51.1)&797.2 (34.9)&921.4 (66.8)&1075.0 (96.6)\\
atlantis&8487.2 (521.1)&976179.7 (8946.6)&927126.9 (16110.0)&940490.4 (19823.0)\\
bankheist&17.4 (3.9)&607.7 (100.9)&934.8 (11.5)&1040.3 (57.6)\\
battlezone&3307.9 (364.7)&29485.5 (949.4)&34422.8 (1031.1)&42040.7 (617.7)\\
beamrider&4326.7 (477.2)&8496.0 (539.6)&9473.1 (806.9)&10797.2 (777.7)\\
berzerk&576.1 (7.9)&767.3 (3.8)&791.3 (7.2)&831.6 (9.1)\\
bowling&26.6 (3.5)&81.7 (7.1)&88.6 (10.7)&97.9 (11.7)\\
boxing&-25.7 (0.8)&52.4 (1.3)&91.6 (6.2)&96.0 (1.6)\\
breakout&21.7 (12.6)&276.8 (14.3)&325.1 (5.0)&370.2 (8.4)\\
centipede&9298.1 (1314.1)&53349.9 (3118.4)&74364.0 (4338.5)&105440.0 (9567.5)\\
choppercommand&565.3 (77.5)&658.6 (218.2)&1272.6 (1610.7)&3118.6 (4640.5)\\
crazyclimber&102366.9 (1650.4)&124196.7 (1221.0)&134382.7 (2569.4)&142029.7 (30.9)\\
demonattack&7502.7 (1316.7)&30118.8 (4122.4)&55757.9 (8478.9)&95685.1 (196.6)\\
doubledunk&-22.9 (0.5)&-21.2 (0.9)&-21.7 (0.5)&-18.4 (2.7)\\
enduro&175.6 (71.2)&1297.6 (63.0)&1714.3 (59.2)&1885.2 (160.4)\\
fishingderby&-90.2 (0.0)&15.1 (0.4)&18.7 (0.9)&20.5 (0.2)\\
freeway&17.9 (0.8)&33.1 (0.1)&33.4 (0.0)&33.5 (0.1)\\
frostbite&660.9 (289.3)&3254.0 (173.8)&3483.5 (225.5)&4023.9 (135.6)\\
gopher&678.2 (171.9)&19015.8 (1765.9)&25183.6 (6676.9)&38405.3 (5558.2)\\
gravitar&174.6 (9.8)&605.9 (81.0)&765.9 (46.7)&975.9 (348.2)\\
hero&3385.2 (199.2)&15007.8 (2516.9)&21247.7 (495.4)&29424.4 (2582.7)\\
icehockey&-15.2 (0.2)&-9.3 (1.1)&-5.6 (0.9)&-5.0 (0.7)\\
jamesbond&248.7 (13.8)&871.3 (193.4)&3642.1 (2146.0)&9270.5 (3173.7)\\
kangaroo&2074.7 (865.0)&10271.2 (75.9)&10526.7 (741.3)&11512.9 (1308.5)\\
krull&2237.3 (177.3)&7705.5 (129.7)&8131.3 (152.2)&8713.5 (297.2)\\
kungfumaster&21452.6 (1944.8)&29539.3 (2798.6)&35743.8 (2124.8)&41563.1 (1753.4)\\
montezumarevenge&0.0 (0.0)&7.0 (1.4)&18.4 (16.2)&39.1 (15.7)\\
mspacman&1447.5 (173.2)&3592.8 (234.5)&4896.7 (269.6)&5508.9 (784.2)\\
namethisgame&2609.4 (339.0)&8001.1 (1213.2)&11231.4 (356.9)&13181.4 (418.4)\\
phoenix&5555.5 (1293.5)&13435.8 (1698.5)&20077.6 (3580.8)&22942.8 (2682.8)\\
pitfall&-43.3 (18.4)&-72.4 (18.2)&-109.4 (60.4)&-177.8 (97.2)\\
pong&-19.7 (1.2)&11.0 (3.7)&15.9 (1.4)&18.2 (1.3)\\
privateeye&131.8 (47.5)&-87.8 (51.8)&-41.7 (91.9)&5513.0 (7933.4)\\
qbert&1032.6 (143.9)&9033.6 (1996.4)&15585.1 (134.6)&20328.0 (2459.4)\\
riverraid&3391.4 (182.8)&12805.4 (194.0)&16284.8 (391.7)&19086.4 (146.5)\\
roadrunner&22377.3 (1106.9)&43889.8 (992.8)&45766.6 (1442.3)&48594.9 (916.6)\\
robotank&6.2 (1.7)&26.9 (3.9)&42.5 (5.1)&61.2 (2.6)\\
seaquest&433.9 (105.2)&3395.6 (259.5)&4072.1 (183.7)&4193.9 (191.7)\\
skiing&-22792.1 (536.1)&-27025.1 (2036.8)&-30319.5 (158.0)&-30591.0 (47.2)\\
solaris&1443.2 (81.3)&1152.0 (189.5)&1349.2 (312.8)&1538.4 (528.1)\\
spaceinvaders&633.2 (31.8)&1039.5 (25.8)&1333.7 (66.2)&1684.0 (118.5)\\
stargunner&1206.4 (84.3)&50894.5 (4573.2)&61879.0 (4897.6)&90312.8 (12514.9)\\
tennis&-23.8 (0.0)&-23.3 (0.8)&-23.8 (0.0)&-23.8 (0.0)\\
timepilot&1191.2 (63.8)&4322.8 (304.1)&6214.5 (546.3)&8156.5 (266.9)\\
tutankham&59.8 (29.6)&52.2 (24.6)&74.6 (17.8)&157.6 (25.5)\\
upndown&7222.4 (409.3)&16852.6 (394.9)&20112.0 (721.4)&25582.3 (1798.3)\\
venture&17.1 (5.0)&9.3 (12.5)&4.1 (5.0)&1.9 (3.3)\\
videopinball&23182.9 (2730.6)&199886.5 (11527.9)&230849.9 (11523.5)&342055.9 (76220.0)\\
wizardofwor&468.1 (48.0)&2901.9 (1149.8)&7466.4 (2461.0)&14566.2 (2521.3)\\
yarsrevenge&9673.8 (188.8)&20966.4 (14486.5)&52049.1 (34059.4)&87772.2 (3939.9)\\
zaxxon&962.5 (223.6)&6574.6 (1478.2)&11183.4 (198.1)&12113.7 (446.0)\\
\bottomrule
\end{tabular}
\caption{Sticky action raw scores for \algnamn{} over all 55 Atari games at various iterations in the training phase as suggested by \cite{machado2018dopamine}. The scores, which are derived from moving averages over 5M frames for each seed, are computed as the mean over all available seeds with one standard deviation included in parentheses.}\label{tab:resultsAllgames}
\end{table}
\clearpage
\section{Mean Learning Curves}
\vspace{1em}
\begin{figure}[h]
    \vskip -1em
    \centering
    \includegraphics[width=\textwidth]{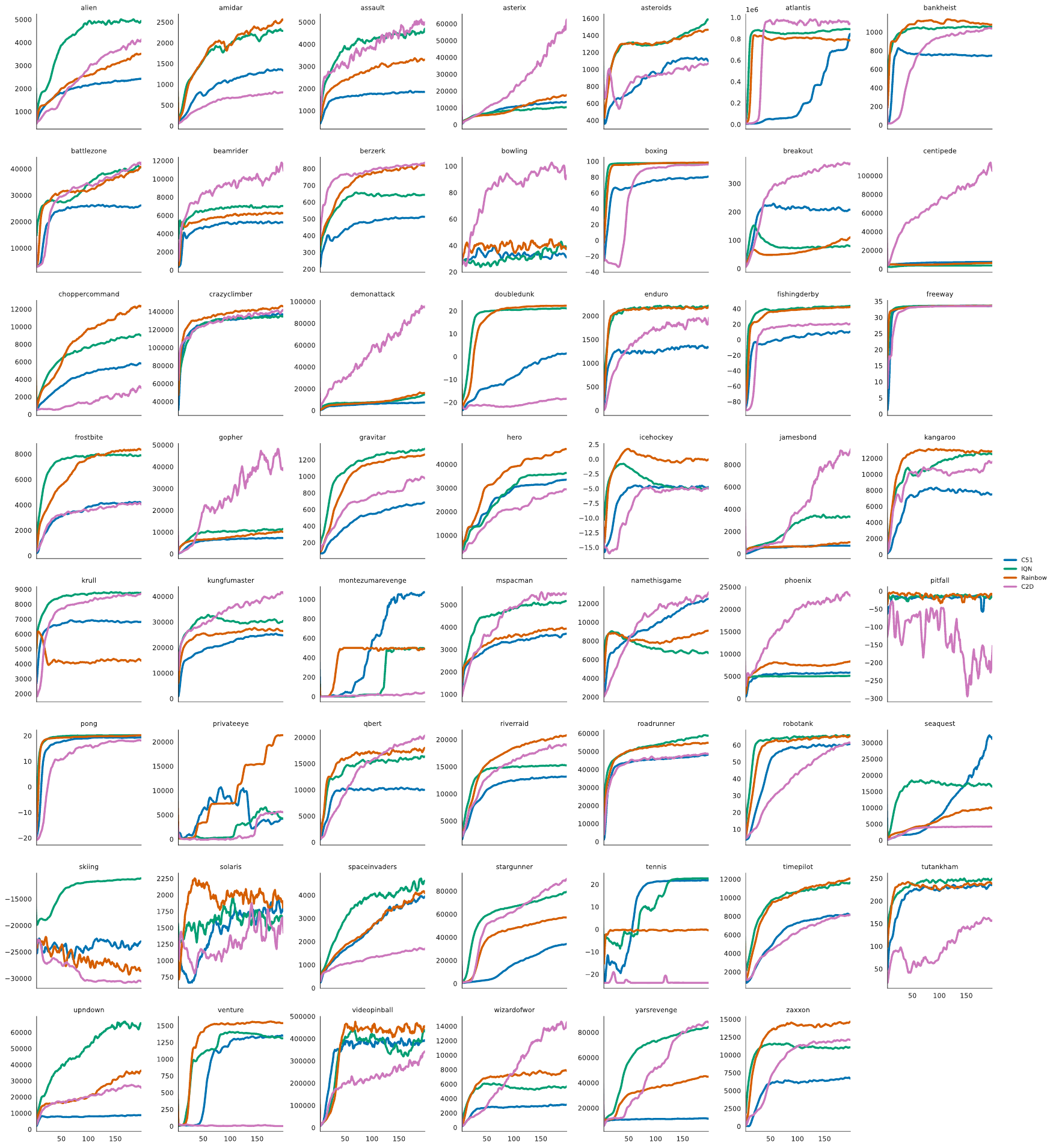}
    \caption{Mean learning curves for all games. Scores are computed by moving averages over 5M frames and the curves are the mean progression over all seeds. \algnamn{} used 3 seeds, and C51, IQN and Rainbow used 5 seeds (Dopamine, 2020).}
    \label{fig:graphs}
\end{figure}
\clearpage

\section{Mean Curves for the Maximal Possible Support}
\vspace{1em}
\begin{figure}[h]
    \vskip -1em
    \centering
    \includegraphics[width=\textwidth]{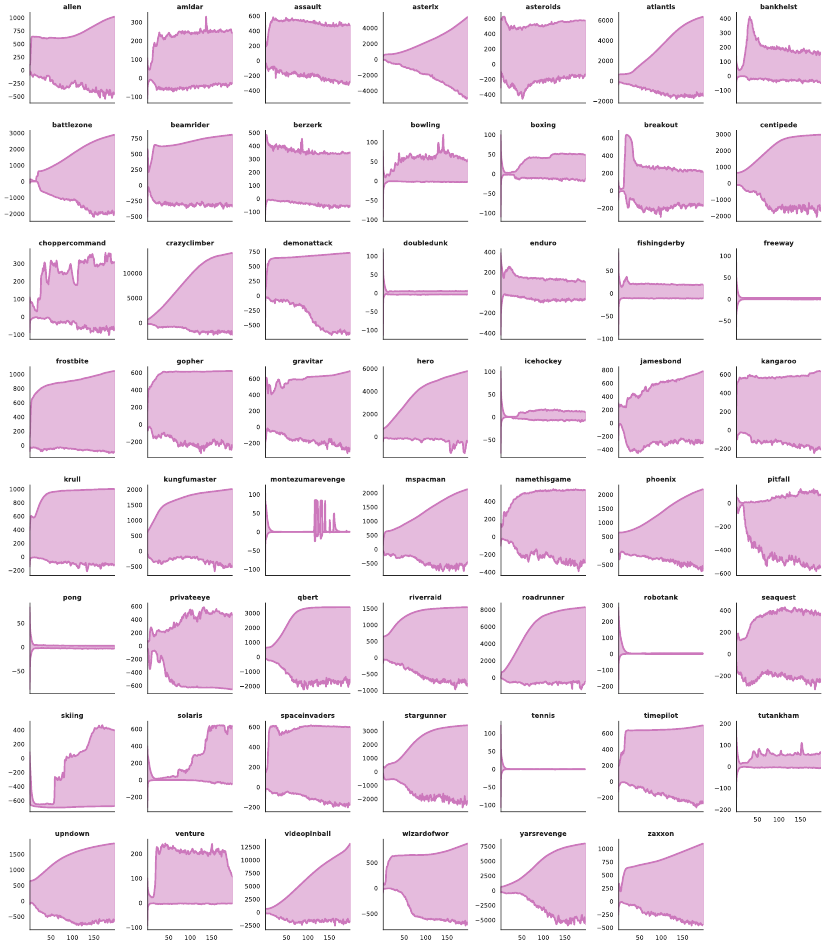}
    \caption{Mean curves for the maximal possible support of C2D over all frames across all 55 games. The larger value is taken as the maximum predicted atom over all actions on the last 1M frames, and the lower as the minimum.}
    \label{fig:supp}
\end{figure}
\clearpage
\section{Sampling Efficiency: Mean and Median}
\begin{figure}[h]
    \centering
    \includegraphics[width=\textwidth]{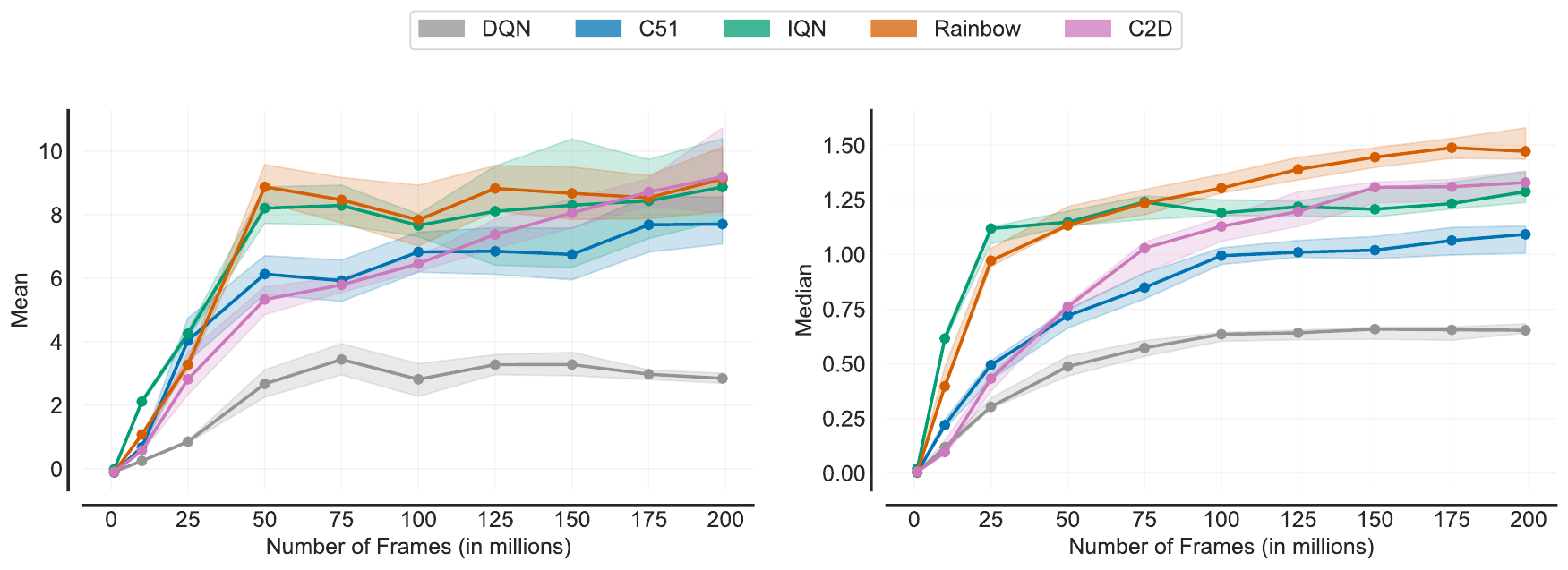}
    \caption{Aggregate mean and median metrics on Atari-200M over 55 games. The metrics are computed in accordance to the performance profiling methods given in \cite{agarwal2021deep}. Dopamine results are computed over 5 runs and \algnamn{} used 3 runs.}
    \label{fig:meanmedian}
\end{figure}
\end{document}